\documentclass{article}

\usepackage{PRIMEarxiv}
\usepackage{tabularx}
\usepackage[utf8]{inputenc}
\usepackage[T1]{fontenc}
\usepackage{lineno}
\usepackage{hyperref}       
\usepackage{url}            
\usepackage{enumitem}
\usepackage{multirow}
\usepackage{graphicx}
\usepackage{subcaption}
\usepackage{colortbl}
\usepackage{mathtools}
\usepackage{tikz}
\usepackage{booktabs}

\definecolor{blue-violet}{rgb}{0.54, 0.17, 0.89}
\definecolor{bluegray}{rgb}{0.4, 0.6, 0.8}
\definecolor{bleudefrance}{rgb}{0.19, 0.55, 0.91}
\definecolor{darkblue}{rgb}{0.0, 0.0, 0.55}
\definecolor{denim}{rgb}{0.08, 0.38, 0.74}
\definecolor{mediumblue}{rgb}{0.0, 0.0, 0.8}
\definecolor{naplesyellow}{rgb}{0.98, 0.85, 0.37}
\definecolor{carminered}{rgb}{1.0, 0.0, 0.22}
\definecolor{crimsonglory}{rgb}{0.75, 0.0, 0.2}
\definecolor{lightgreen}{rgb}{0.56, 0.93, 0.56}
\definecolor{lightapricot}{rgb}{0.99, 0.84, 0.69}
\definecolor{pastelred}{rgb}{1.0, 0.41, 0.38}
\definecolor{periwinkle}{rgb}{0.8, 0.8, 1.0}
\definecolor{xgreen}{rgb}{0.88, 0.95, 0.83}
\definecolor{uared}{rgb}{0.85, 0.0, 0.3}
\definecolor{debianred}{rgb}{0.84, 0.04, 0.33}
\definecolor{red(ncs)}{rgb}{0.77, 0.01, 0.2}
\definecolor{blush}{rgb}{0.87, 0.36, 0.51}
\definecolor{non-photoblue}{rgb}{0.64, 0.87, 0.93}
\definecolor{lightcornflowerblue}{rgb}{0.6, 0.81, 0.93}
\definecolor{cadmiumgreen}{rgb}{0.0, 0.42, 0.24}
\definecolor{dollarbill}{rgb}{0.52, 0.73, 0.4}
\definecolor{darkgreen}{rgb}{0.0, 0.2, 0.13}
\newcommand\chqsumm{\texttt{CHQ-Summ}}

\newcommand\prophetnet{ProphetNet}
\newcommand\meqsum{\textsc{MeQSum}}
\title{\chqsumm{}: A Dataset for Consumer Healthcare Question Summarization}

\author{
  Shweta Yadav\\
  Department of Computer Science, University of Illinois at Chicago\\
  Chicago, IL, USA \\
  \texttt{shwetay@uic.edu}
  \AND
 Deepak Gupta, and Dina Demner-Fushman \\
  Lister Hill National Center for Biomedical Communications \\
  National Library of Medicine, National Institutes of Health \\
  Bethesda, MD, USA\\
  \texttt{\{firstname.lastname\}@nih.gov} 
}

\begin{document}
\maketitle

\begin{abstract}
The quest for seeking health information has swamped the web with consumers' health-related questions. Generally, consumers use overly descriptive and peripheral information to express their medical condition or other healthcare needs, contributing to the challenges of natural language understanding. One way to address this challenge is to summarize the questions and distill the key information of the original question. 
Recently, large-scale datasets have significantly propelled the development of several summarization tasks, such as multi-document summarization and dialogue summarization. However, a lack of domain-expert annotated dataset for the consumer healthcare questions summarization task inhibits the development of an efficient summarization system.
To address this issue, we introduce a new dataset \chqsumm{}{} that contains $1507$ domain-expert annotated consumer health questions and corresponding summaries. The dataset is derived from the community question answering forum and therefore provides a valuable resource for understanding consumer health-related posts on social media.
We benchmark the dataset on multiple state-of-the-art summarization models to show the effectiveness of the dataset.

\end{abstract}

\flushbottom
\maketitle



\section{Introduction}
Healthcare consumers often query the web to find a quick and reliable answer to their healthcare information needs. On average, six million people in the United States alone seek health-related information on the Internet every day \cite{foxwebsite}. One way to facilitate such information-seeking activities is to build a natural language question-answering (QA) system that can extract precise answers from a myriad of health-related information sources. Though existing search engines respond to the general health-related queries to some extent, users often reach out to specialized medical websites or online health communities to seek personalized, high-quality, and trustworthy answers to their complex health questions.
The overly descriptive nature of the questions that contain too much peripheral information brings an additional challenge to the task of automatic analysis and understanding of the questions. Often, such peripheral details are not required to obtain relevant answers, and their removal can lead to significant improvement in QA performance. Therefore, there is a need to develop  {\em automatic question simplification/summarization techniques} before retrieving answers to the questions.

Automatic text summarization is a non-trivial task in natural language processing that aims to generate human-readable, concise text containing salient information of the original document.
The recent development in large-scale neural language models \cite{devlin-etal-2019-bert,raffel2020exploring} has led to
significant performance improvement on several summary generation tasks ( called abstractive summarization), partially due to the availability of large-scale human-annotated training data. The majority of the current summarization datasets are either based on the news articles (e.g., CNN/Dailymail\cite{nallapati-etal-2016-abstractive} and Multi-News\cite{fabbri2019multi} datasets where headlines are treated as summaries) or the scientific literature ( e.g., PubMed\cite{cohan-etal-2018-discourse}, BioASQ\cite{tsatsaronis2015overview} datasets where abstracts of the articles serve as summaries).
While significant efforts have been made in the open-domain summarization, there are only a few works\cite{yadav2022question,abacha2019summarization, yadav2021nlm,yadav2021reinforcement,sarrouti2021nlm} in summarizing the CHQs. This is partially due to the lack of availability of the human-annotated training datasets. 
Recently, a manually annotated dataset, \meqsum{} \cite{abacha2019summarization} was created that consists of $1000$ CHQs and their corresponding summaries. A related dataset, MEDIQA-AnS  \cite{savery2020question} focuses on answer summarization for consumer health question answering. While these datasets enabled the research in question answering, the amount of training data is still relatively small, hindering the progress in developing an accurate and efficient CHQ summarization system.

Towards this, we introduce a new CHQ summarization dataset: \chqsumm{}{} that consists of $1507$ domain-expert annotated question-summary pairs from the Yahoo community question answering forum. We advanced the existing dataset in two ways: \textbf{(i)} the dataset is created from the community question answering forum having a more diverse set of users' questions, and \textbf{(ii)} our \chqsumm{}{} dataset contains additional annotations about question focus and question type of the original question. The question focus is the main entity of the question and the question type is the aspect of interest about the focus (\textit{c.f.} Figure \ref{fig1}).

\section{Methods}
\subsection{Dataset Creation} We utilized the Yahoo! Answers L6 corpus\footnote{\url{https://webscope.sandbox.yahoo.com/catalog.php?datatype=l&did=11}} that provides community question answering threads containing users' questions on multiple diverse topics and the answers submitted by other users. 
The corpus consists of $4.4$ million question-answer threads with additional metadata information such as the question category, question sub-category, question language, and best answer. Each of the questions has a question title and question content. The question title is the subject line of the original question, and the question content describes the original question with the full description. Since our goal is focused on the consumer healthcare domain, we selected the questions from the ``\textit{Healthcare}" question category. However, to ensure that there are \textbf{(i)} no false positives, and \textbf{(ii)} the questions are from diverse health categories, we devised multiple heuristic-based filtering strategies to filter out irrelevant questions, discussed as follows: 
\begin{figure}
    \centering
    \includegraphics[width=0.8\linewidth]{images/Fig5-1.png}
    \caption{Sample consumer healthcare question and reference summary with the annotated question focus and question type.}
    \label{fig1}
\end{figure}
\begin{itemize}[noitemsep]
    \item \textbf{Medical Entities Identification:} The very first stage of filtration is to recognize the medical entities present in the question content and title. 
    We utilized Stanza \cite{zhang2021biomedical} Biomedical and Clinical model 
    for identifying medical entities in the question title and content. 
    %
    \item \textbf{Candidate Question Identification and Selection:} 
    In the second stage, we select the question threads having at least one medical entity present in either question title or question content. We collected $22,257$ question threads from Yahoo! Answers corpus through this process.

\item \textbf{Removing Low Content Question Threads:} In the final stage of filtration, to preserve the quality of our dataset, we further filter out the questions that are short and do not require summarization. 

 Specifically, we removed the question threads with a question content having less than $10$ words. Later, we concatenate the question title and content to form the original question and ask annotators to formulate the summary of the question manually. We call this manually generated question-summary pair \chqsumm{}{}. These pairs constitute the consumer health question summarization dataset. The schematic workflow of the \chqsumm{} dataset creation is depicted in Figure \ref{approach}.

\end{itemize}

\begin{figure}
    \centering
    \includegraphics[width=13cm, height=10cm]{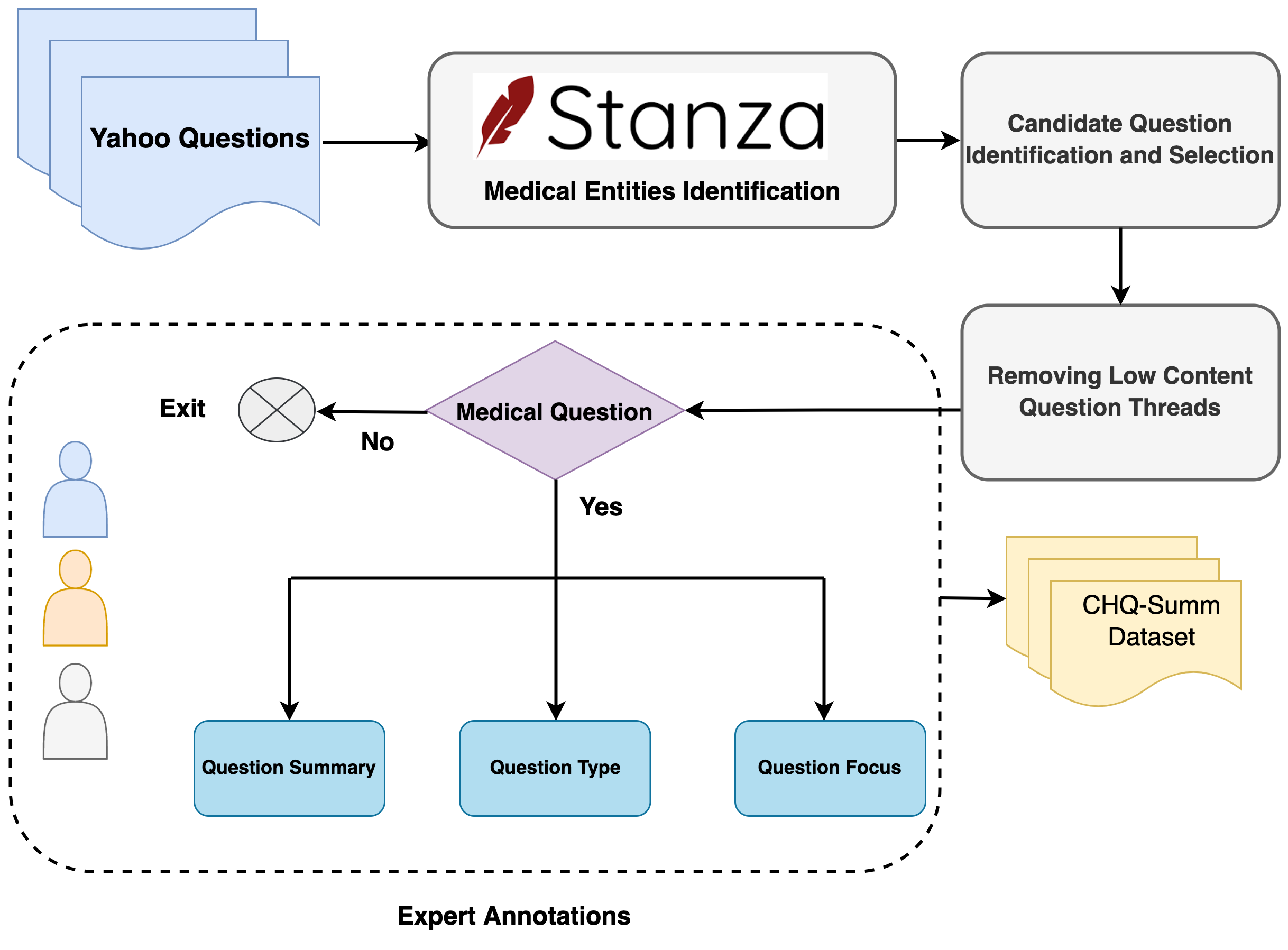}
    \caption{The schematic workflow of the \chqsumm{} dataset creation.}
    \label{approach}
\end{figure}

\subsection{Expert Annotations} A team of six annotators ($4$ experts in medical informatics and $2$ experts in medical informatics and medicine) independently annotated  parts of the \chqsumm{}{} dataset. Each annotator was provided with the annotation guidelines and the annotation interface to summarize the CHQs. Given the original question, the annotators were instructed to:
\begin{enumerate}
    \item \textbf{Identify the valid medical question:} A question can be categorized as a valid medical question if it belongs to the following categories: 
    \begin{enumerate}[noitemsep]
        \item Diseases \& conditions, including symptoms
        \item Drugs \& treatments
        \item Medical tests or medical diagnostic \& therapeutic procedures
        \item Other relevant medical topics 
    \end{enumerate}
    \item \textbf{Formulate an abstractive summary:} The abstractive summary of the given question is formulated by keeping the minimum and key information required to answer the question. 
    \item \textbf{Identify the question focus:} 
    Given the effectiveness of the question focus for the consumer health question summarization task, we ask the annotators to annotate each question with the focus term. The question focus is the named entity that is the central theme of the question.
   Generally, a question has a single focus, but for some questions which are about the relation between two drugs or diseases, multiple question foci are allowed. For example, consider Fig. \ref{fig1} where the question focus is \textit{`cataract'}.

    \item \textbf{Identify the question type:} Question type represents some aspect of the question focus that triggered the question and needs to be answered, e.g., in Fig. \ref{fig1} the expecting mother is asking if her child will be susceptible to cataracts more than an average baby. 
    Existing studies\cite{yadav2022question} show that question-type information can guide the model to generate more factually correct summaries, therefore, we annotated each question with the appropriate question types.
    Specifically, we instructed the annotators to classify each question into any of the $33$ question type categories obtained from the previous study \cite{kilicoglu2018semantic}.
    The annotators were expected to select all the questions types that could be inferred from the given question. For example, in the Fig. \ref{fig1}, the question type is \textit{`Susceptibility'}.

\end{enumerate}

\subsection{Annotators and Inter-annotation Agreement} 
We began the annotation process by developing the annotation guidelines and interface. In the first round, a set of $3$ annotators with expertise in medical informatics and medicine annotated the same collection of $150$ questions to compute the inter-annotation agreement. In the second round, $3$ more annotators with medical informatics background were included in the annotation task. For the final annotation, a total of $6$ annotators independently annotated the collection of $500$ questions, out of which $30$ questions were common among all the annotators. We computed the inter-annotator agreement (IAA) on the $30$ common question. 

For computing the IAA for the summarization task, we used the ROUGE-L score. In the case of question-focus and question-type identification tasks, we used F1-score to measure the IAA. We have reported the average ROUGE-L and F1-score amongst all the annotator pairs in Table-\ref{tab:IAA}.
The average IAA for question focus and question types was found to be $83.22$ and $85.44$ (substantial agreement), respectively, on $30$ common questions. On the summarization task, we observed that even though the summary was semantically correct, the average IAA was only $52.8$ (moderate agreement). This was because the summary created by the annotators was diverse and did not follow the same word order, even though they all were semantically correct. Since ROUGE-L does not take into account the semantic similarity between two summaries, the ROUGE-L score was not high among annotators. For example, consider Figure-\ref{fig1}, where both variations of the question ask the same thing but don't have too many words in common.

\begin{table}[]
\centering
\resizebox{0.7\textwidth}{!}{%
\begin{tabular}{c|c|c|c}
\hline
\textbf{Samples/Annotators} &
  \textbf{\begin{tabular}[c]{@{}c@{}}Question-focus\\  (F1-score)\end{tabular}} &
  \textbf{\begin{tabular}[c]{@{}c@{}}Question-type\\ (F1-score)\end{tabular}} &
  \textbf{\begin{tabular}[c]{@{}c@{}}Question-summary\\  (ROUGE-L)\end{tabular}} \\ \hline
150/3 &
  85.84 &
  86.28 &
  53.23 \\ 
30/6 &
  83.22 &
  85.44 &
  52.8 \\ \hline
\end{tabular}%
}
\caption{Average inter-annotator agreement score among all the annotators on question-focus recognition, question-type identification, and question-summary generation task.}
\label{tab:IAA}
\end{table}

\begin{figure}[h]
\centering
\begin{subfigure}{.5\textwidth}
\centering
\includegraphics[scale=0.5]{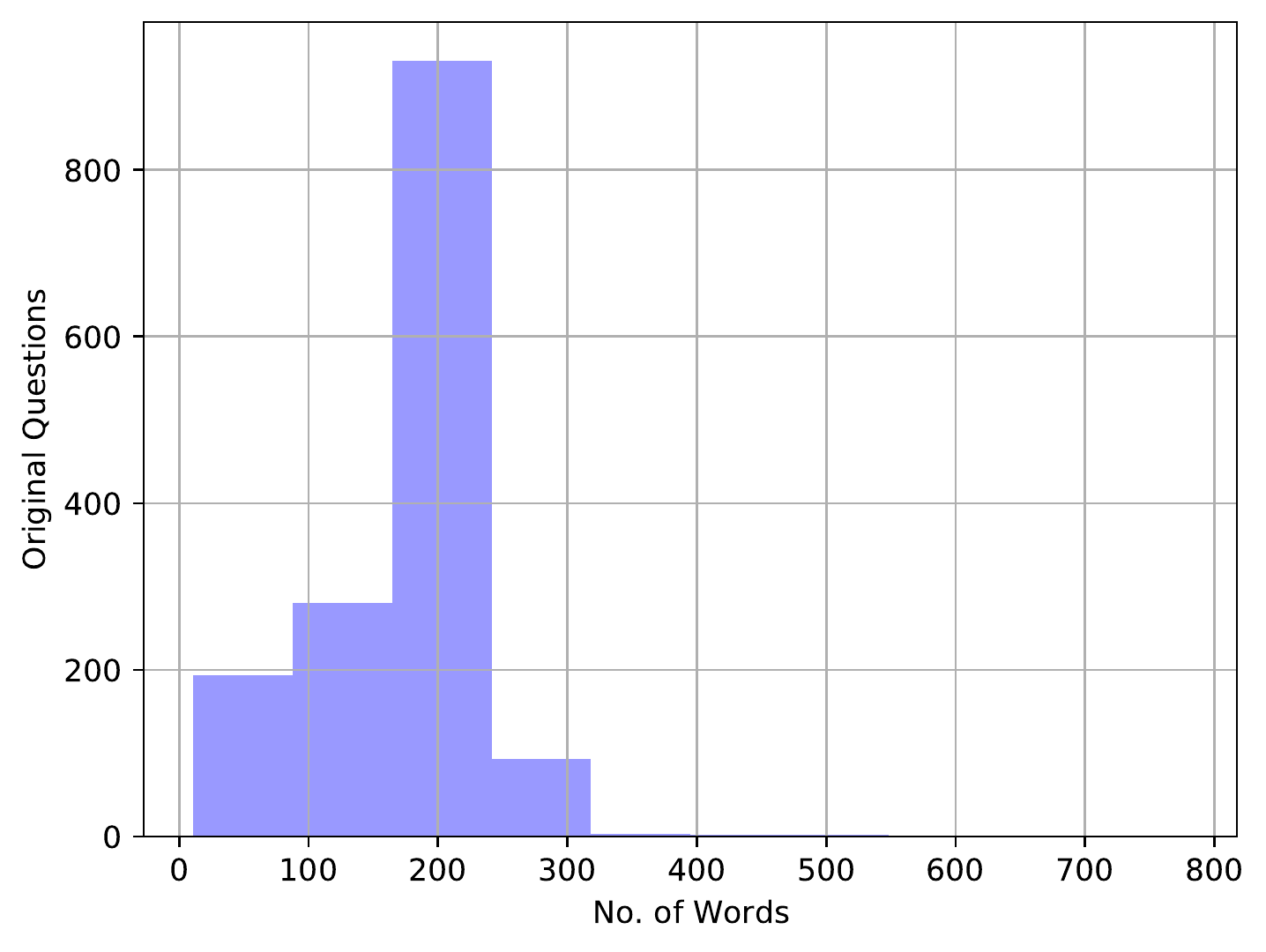}
\caption{}
\end{subfigure}%
\begin{subfigure}{.5\textwidth}
\centering
    \includegraphics[scale=0.5]{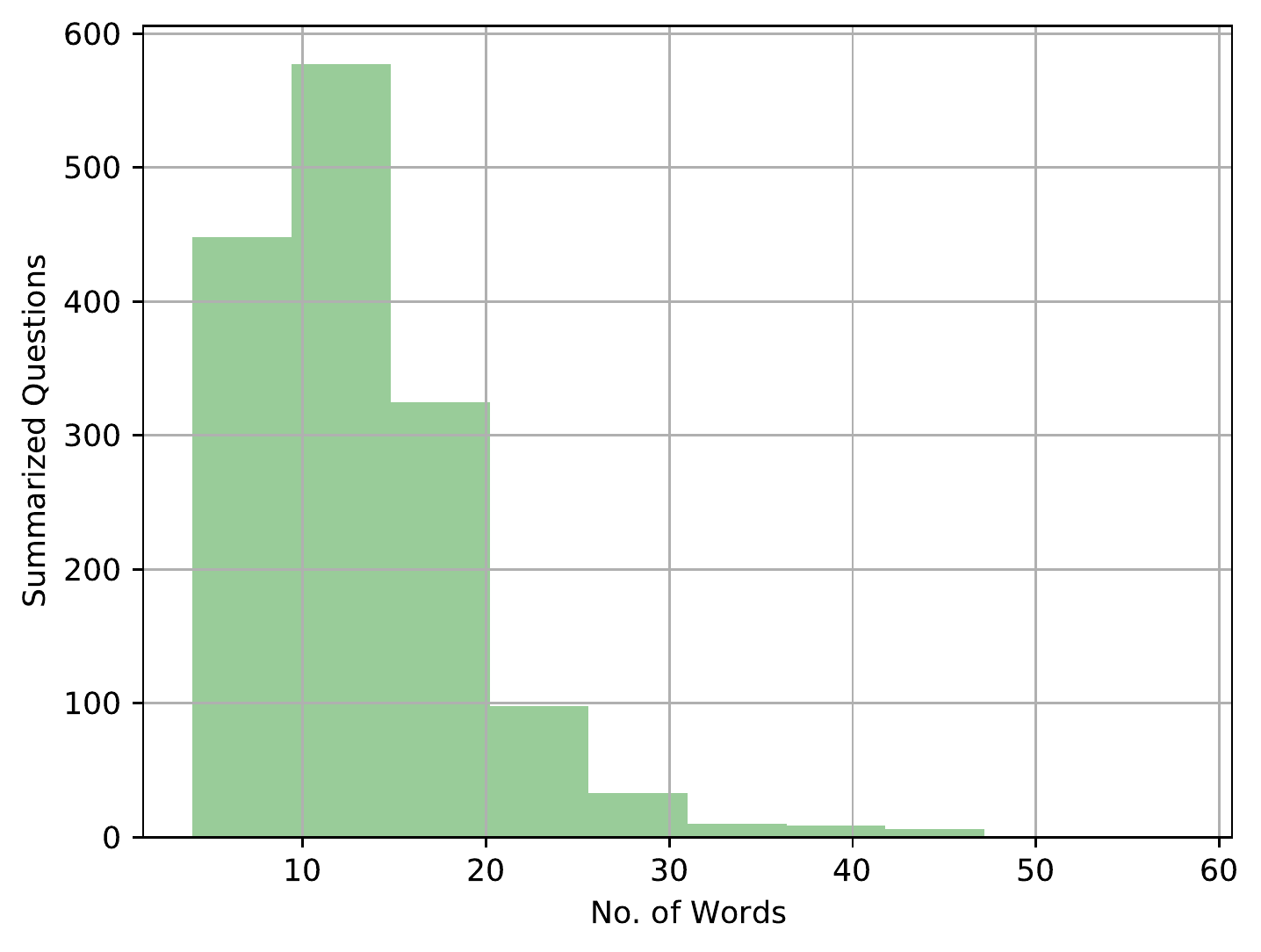}
    \caption{}
\end{subfigure}%
    \caption{Distribution of words in original questions \textbf{(a)} and summarized questions \textbf{(b)} in \chqsumm{} dataset.}
\label{fig:word_distribution}
\end{figure}

\begin{figure*}[t]
\begin{minipage}{.6\textwidth}
\centering
\resizebox{\textwidth}{!}{%
\begin{tabular}{lc|c|c|c|c}
\hline \hline
\multicolumn{2}{c|}{\textbf{Statistics}} &
  \textbf{\begin{tabular}[c]{@{}c@{}}Train\\ (Sentence/Word)\end{tabular}} &
  \textbf{\begin{tabular}[c]{@{}c@{}}Test\\ (Sentence/Word)\end{tabular}} &
  \textbf{\begin{tabular}[c]{@{}c@{}}Validation\\ (Sentence/Word)\end{tabular}} &
  \textbf{\begin{tabular}[c]{@{}c@{}}Total\\ (Sentence/Word)\end{tabular}} \\ \hline
\multicolumn{1}{l|}{\multirow{5}{*}{\textbf{Question}}} & \textbf{Samples}            & 1000         & 400         & 107         & 1507         \\ 
\multicolumn{1}{l|}{}                                   & \textbf{Minimum}            & 1/11         & 2/14        & 2/12        & 1/11         \\ 
\multicolumn{1}{l|}{}                                   & \textbf{Maximum}            & 32/779       & 35/628      & 23/260      & 35/779       \\ 
\multicolumn{1}{l|}{}                                   & \textbf{Mean}               & 10.04/177.21 & 9.99/177.79 & 9.94/168.58 & 10.02/176.75 \\  
\multicolumn{1}{l|}{}                                   & \textbf{Standard Deviation} & 4.54/63.20   & 4.69/68.64  & 4.59/64.01  & 4.58/64.74   \\ \hline
\multicolumn{1}{l|}{\multirow{5}{*}{\textbf{Summary}}}  & \textbf{Samples}            & 1000         & 400         & 107         & 1507         \\
\multicolumn{1}{l|}{}                                   & \textbf{Minimum}            & 1/4          & 1/4         & 1/4         & 1/4          \\
\multicolumn{1}{l|}{}                                   & \textbf{Maximum}            & 2/58         & 2/47        & 2/45        & 2/58         \\ 
\multicolumn{1}{l|}{}                                   & \textbf{Mean}               & 1.02/13      & 1.03/13.62  & 1.02/13.88  & 1.02/13.23   \\  
\multicolumn{1}{l|}{}                                   & \textbf{Standard Deviation} & 0.15/6.03    & 0.18/6.09   & 0.16/7.16   & 0.16/6.14    \\ \hline
  \hline
\end{tabular}%
}
\captionof{table}{Dataset statistics for consumer health question and its corresponding summary in train, test and validation set.}
\label{tab:question-summary-stats}
\end{minipage}
\centering
\quad
\begin{minipage}{.35\textwidth}
\centering
\resizebox{\textwidth}{!}{%
\begin{tabular}{lc|c|c|c|c}
\hline \hline
\multicolumn{2}{c|}{\textbf{Statistics}} &
  \textbf{\begin{tabular}[c]{@{}c@{}}Train\end{tabular}} &
  \textbf{\begin{tabular}[c]{@{}c@{}}Test\end{tabular}} &
  \textbf{\begin{tabular}[c]{@{}c@{}}Validation\end{tabular}} &
  \textbf{\begin{tabular}[c]{@{}c@{}}Total\end{tabular}} \\ \hline
\multicolumn{1}{l|}{\multirow{4}{*}{\textbf{Question Type}}}  & \textbf{Minimum} & 1              & 1             & 1                   & 1              \\ 
\multicolumn{1}{l|}{}                  & \textbf{Maximum}            & 4    & 3    & 3    & 4    \\ 
\multicolumn{1}{l|}{}                  & \textbf{Mean}               & 1.17 & 1.19 & 1.14 & 1.18 \\ 
\multicolumn{1}{l|}{}                  & \textbf{Standard Deviation} & 0.48 & 0.49 & 0.44 & 0.48 \\ \hline
\multicolumn{1}{l|}{\multirow{4}{*}{\textbf{Question Focus}}} & \textbf{Minimum} & 1              & 1             & 1                   & 1              \\ 
\multicolumn{1}{l|}{}                  & \textbf{Maximum}            & 20   & 21   & 9    & 21   \\ 
\multicolumn{1}{l|}{}                  & \textbf{Mean}               & 1.93 & 2.02 & 1.82 & 1.95 \\ 
\multicolumn{1}{l|}{}                  & \textbf{Standard Deviation} & 1.63 & 1.79 & 1.33 & 1.65 \\ \hline  \hline
\end{tabular}%
}
\captionof{table}{Dataset Statistics about question-type and question-focus in train, test and validation set. }
\label{tab:focus-type-stats}
\end{minipage}
\end{figure*}

\begin{figure}[]
\centering
\begin{subfigure}{.48\textwidth}
\centering
  \includegraphics[scale=0.4]{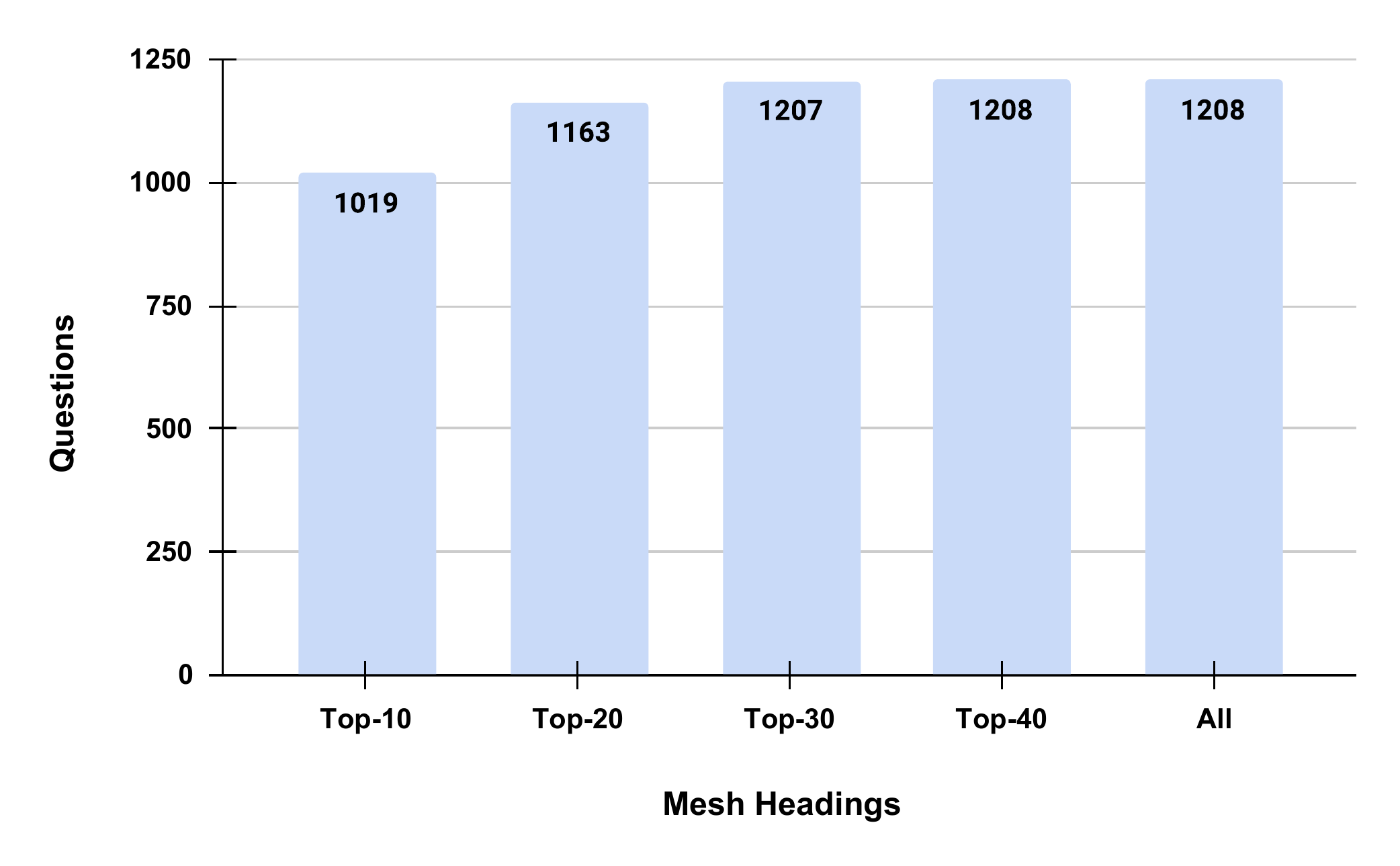}
\caption{}
\end{subfigure}%
\begin{subfigure}{.48\textwidth}
\centering
  \includegraphics[scale=0.4]{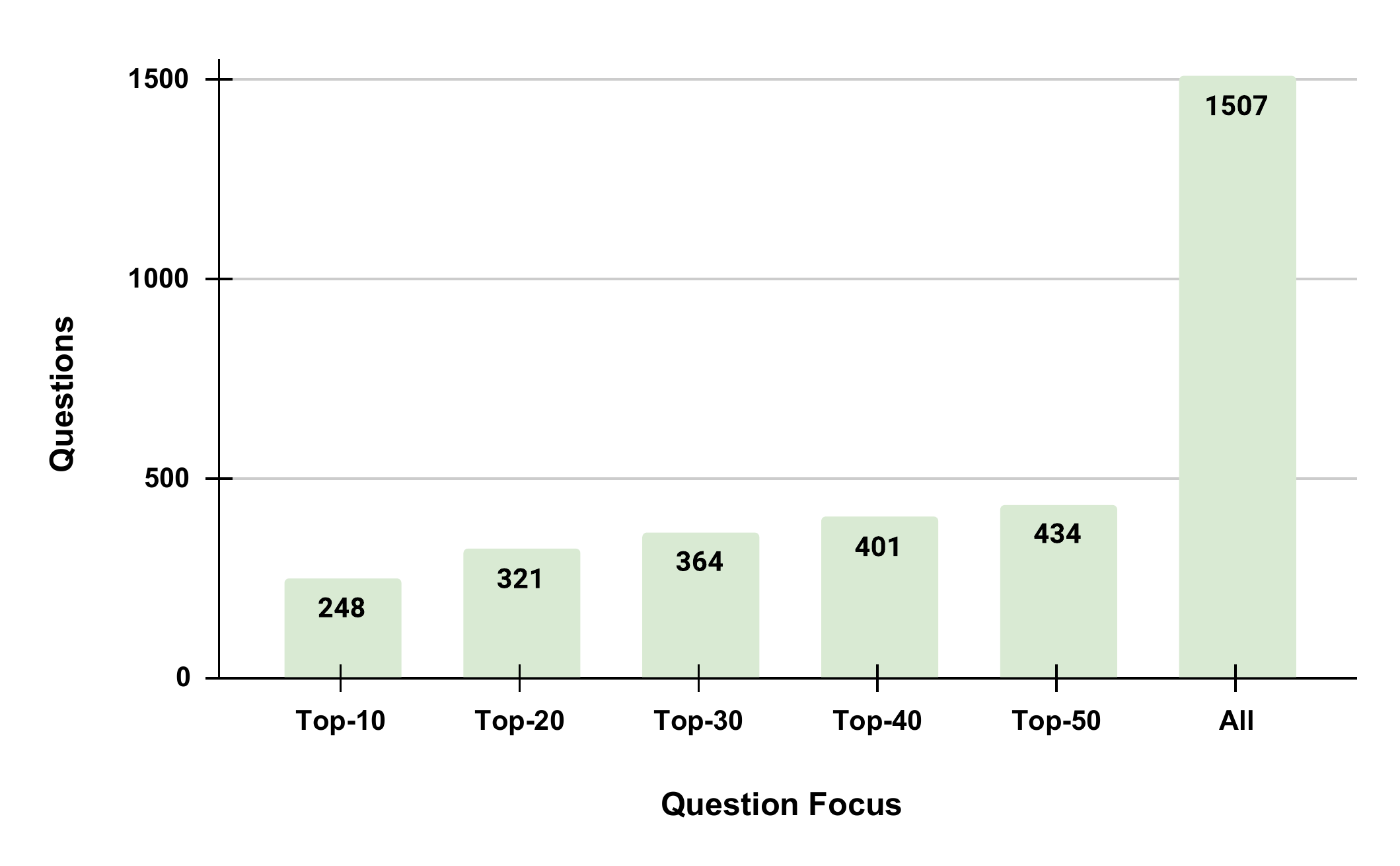}
    \caption{}
\end{subfigure}%
\caption{Coverage of the MeSH headings (\textit{left}) and question focus terms (\textit{right}) on the \chqsumm{} dataset. }
\label{fig:mesh_headings_top_focus}
\end{figure}

\section{Dataset Analysis}
 The \chqsumm{} dataset consists of the $1,507$ original questions and the corresponding summarized questions, question focus and question types. To benchmark the dataset, we split it into training, validation and test set of $1000$, $400$ and $107$ samples respectively. Table \ref{tab:question-summary-stats} and Table \ref{tab:focus-type-stats} provide the detailed statistics of the dataset.

The majority of original questions in \chqsumm{} dataset have $200$ words. While, most of our human generated summaries have $15$ words (\textit{cf.} Fig \ref{fig:word_distribution}). 
This shows that in the community question answering forum, users prefer expressing their healthcare information needs in an overly descriptive manner with several peripheral details rather than formulating succinct queries that demand more cognitive effort. We also analyzed the distribution of the question focus and question type of the \chqsumm{} dataset. To understand the concept associated with each question focus, we map each focus term to the Medical Subject Headings\footnote{\url{https://www.nlm.nih.gov/mesh/meshhome.html}}. 

\paragraph{Finding Medical Subject Headings (MeSH):} 
The Medical Subject Headings is a thesaurus developed by the National Library of Medicine, and it is used for indexing, cataloging, and searching for biomedical and health-related information and documents. Since biomedical literature is indexed with MeSH, it is appropriate to use MeSH to search the literature to find the relevant and reliable answers to consumer healthcare questions.
Toward this, we have provided the MeSH headings for the question focus present in each question of the \chqsumm{} dataset.
We downloaded and pre-processed the Unified Medical Language System\textregistered (UMLS) \cite{bodenreider2004unified} Metathesaurus  from the official UMLS site\footnote{\url{https://www.nlm.nih.gov/research/umls/licensedcontent/umlsknowledgesources.html}}. In particular, we utilized the \texttt{MRCONSO.RRF} Metathesaurus file that contains the concepts id, names, codes, etc., from multiple source vocabularies. To map the MeSH headings associated with the question focus, we downloaded and pre-processed the MeSH Descriptor Data 2021 from the official MeSH website\footnote{\url{https://www.nlm.nih.gov/databases/download/mesh.html}}. In the pre-processing step, we build a MeSH dictionary where the MeSH descriptor is stored as  key, and the MeSH tree numbers are stored as the value. 

Given a question focus, we follow the following steps to obtain the MeSH heading:

\begin{enumerate}[noitemsep]
    \item We utilized the MetaMap \cite{demner2017metamap,aronson2010overview} tool with the MeSH vocabulary that provides the concept unique identifier (CUI) of the mapping concepts.  
     \item The descriptor unique identifier (DUI) for each mapping CUIs are extracted from  \texttt{MRCONSO.RRF} Metathesaurus. While searching the DUI, we restrict ourselves to only extracting the DUIs for which the language of the term is English (ENG) and the source vocabulary is MeSH (MSH). 
    \item Finally, we used the processed MeSH dictionary to retrieve the associated MeSH Tree numbers and mapped them to their top tree numbers in MeSH Tree.
\end{enumerate}
\begin{figure}[]
\centering
\begin{subfigure}{.5\textwidth}
\centering
  \includegraphics[scale=0.55]{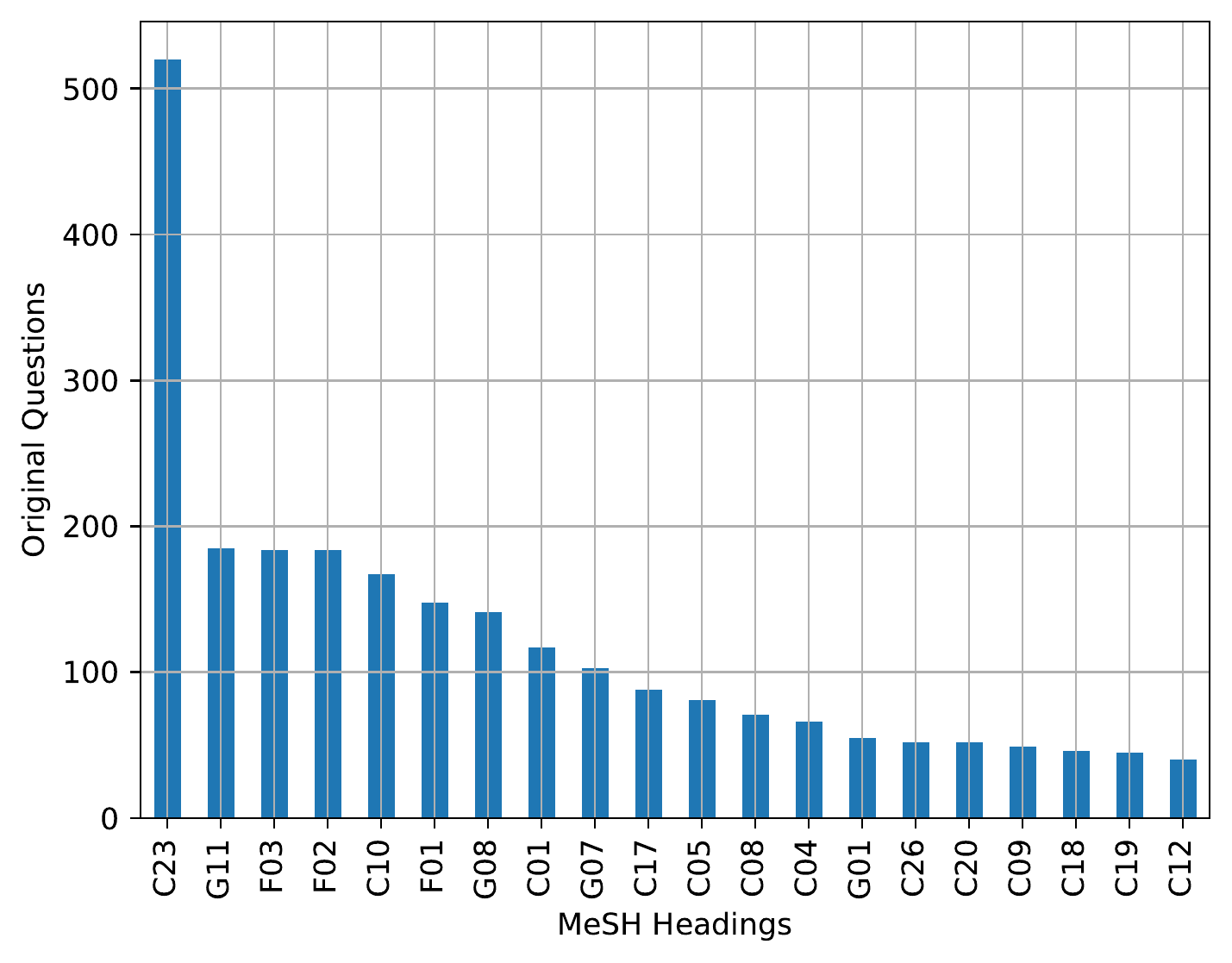}
\caption{}
 \label{fig:mesh-categories}
\end{subfigure}%
\begin{subfigure}{.5\textwidth}
\centering
  \includegraphics[scale=0.55]{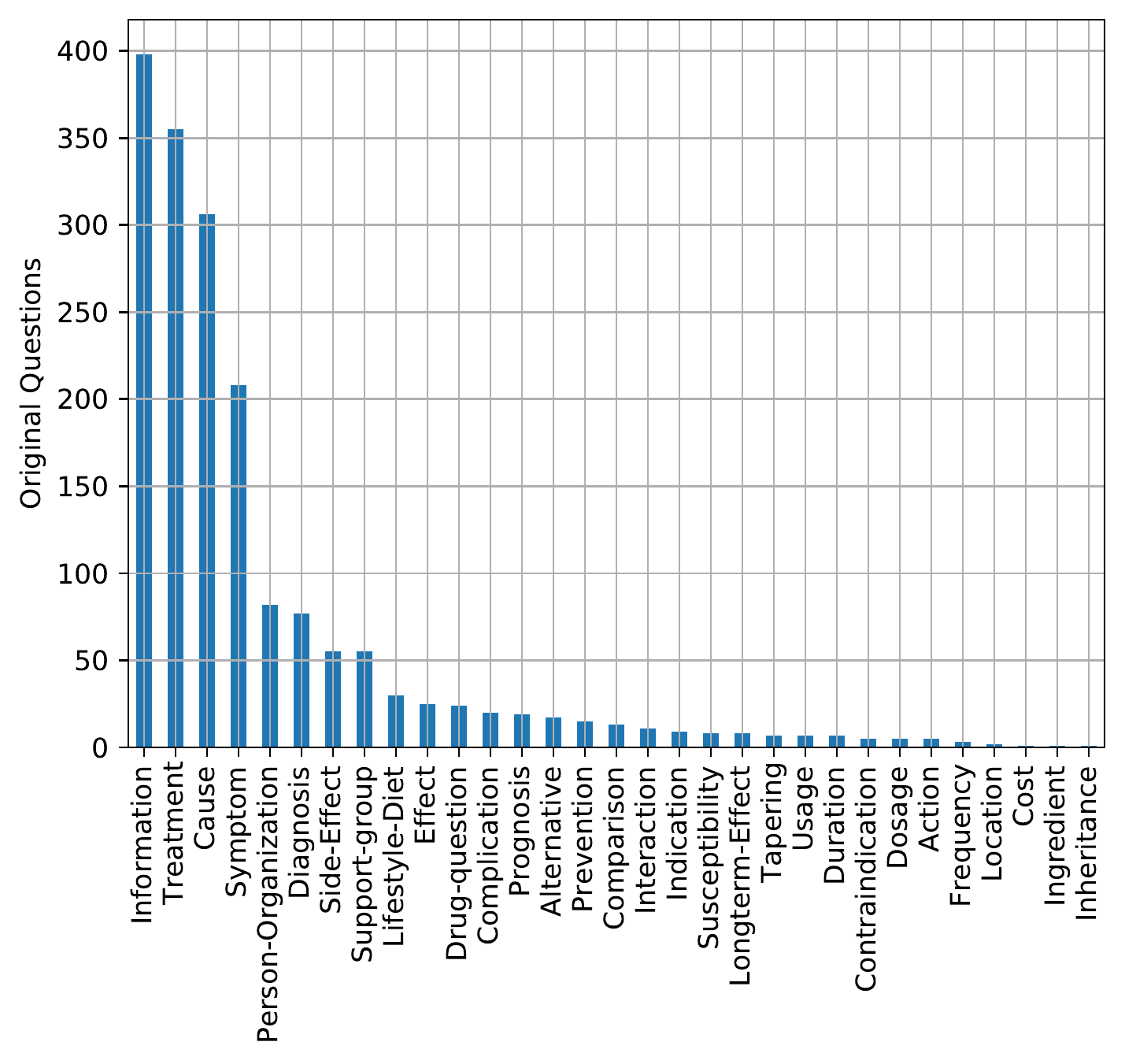}
    \caption{}
\end{subfigure}%
    \caption{Distribution of the MeSH terms corresponds to the question-focus terms annotated in the original questions of the \chqsumm{} dataset (\textit{left}). The \textit{right} side of the Figure shows the distribution of the different question-types annotated in the \chqsumm{} dataset.}
\label{fig:focus_and_type_distribution}
 \end{figure}
With the above approach to find MeSH terms, we were able to obtain the top heading (associated with the MeSH Tree numbers\footnote{Please refer to  \url{https://meshb-prev.nlm.nih.gov/treeView} for MeSH categories.}) for the question focus present in $1,208$ questions out of total $1,507$ questions in \chqsumm{} dataset. The \chqsumm{} dataset contains a total of $1788$ distinct question focuses and with MetaMap we successfully mapped $1141$ question focuses to MeSH terms. The distribution of top $20$ MeSH terms corresponding to the question-focus terms is provided in Fig-\ref{fig:focus_and_type_distribution} (a). The mapping shows that questions on \textit{``pathological conditions signs and symptoms''} (C23) are most frequent, followed by ``\textit{musculoskeletal and neural physiological phenomena }'' (G11) and ``\textit{mental disorders}'' (F03). The top $20$ MeSH mappings contribute to $96$\% of the total annotated dataset (\textit{cf.} Fig. \ref{fig:mesh_headings_top_focus} (a)). We also analyze the coverage (\textit{cf.} Fig. \ref{fig:mesh_headings_top_focus} (b)) of the most frequent $k$ question focuses and found that top-10, top-20 and top-30 question focuses cover $248$, $321$ and $364$ questions respectively. We noticed that most of the questions in the \chqsumm{} dataset have two focus terms as shown in (cf. Fig. \ref{fig:focus_term_and_top_20_focus} (a)). We have provided the distribution of top-20 question focus term in the (\textit{cf.} Fig. \ref{fig:focus_term_and_top_20_focus} (b)). This shows that the \chqsumm{} contains diverse set of questions covering different question focuses.
We also conducted analysis on the annotated question types. The statistics regarding question type are provided in Fig. \ref{fig:focus_and_type_distribution} (b). The distribution shows that consumers frequently seek general information regarding their healthcare needs. As such,\textit{`information'} question type is the most common question type in \chqsumm{}. Other common question types are `\textit{treatment}', `\textit{cause}', and `\textit{symptoms}'.

\begin{figure}[h]
\centering
\begin{subfigure}{.5\textwidth}
\centering
  \includegraphics[scale=0.55]{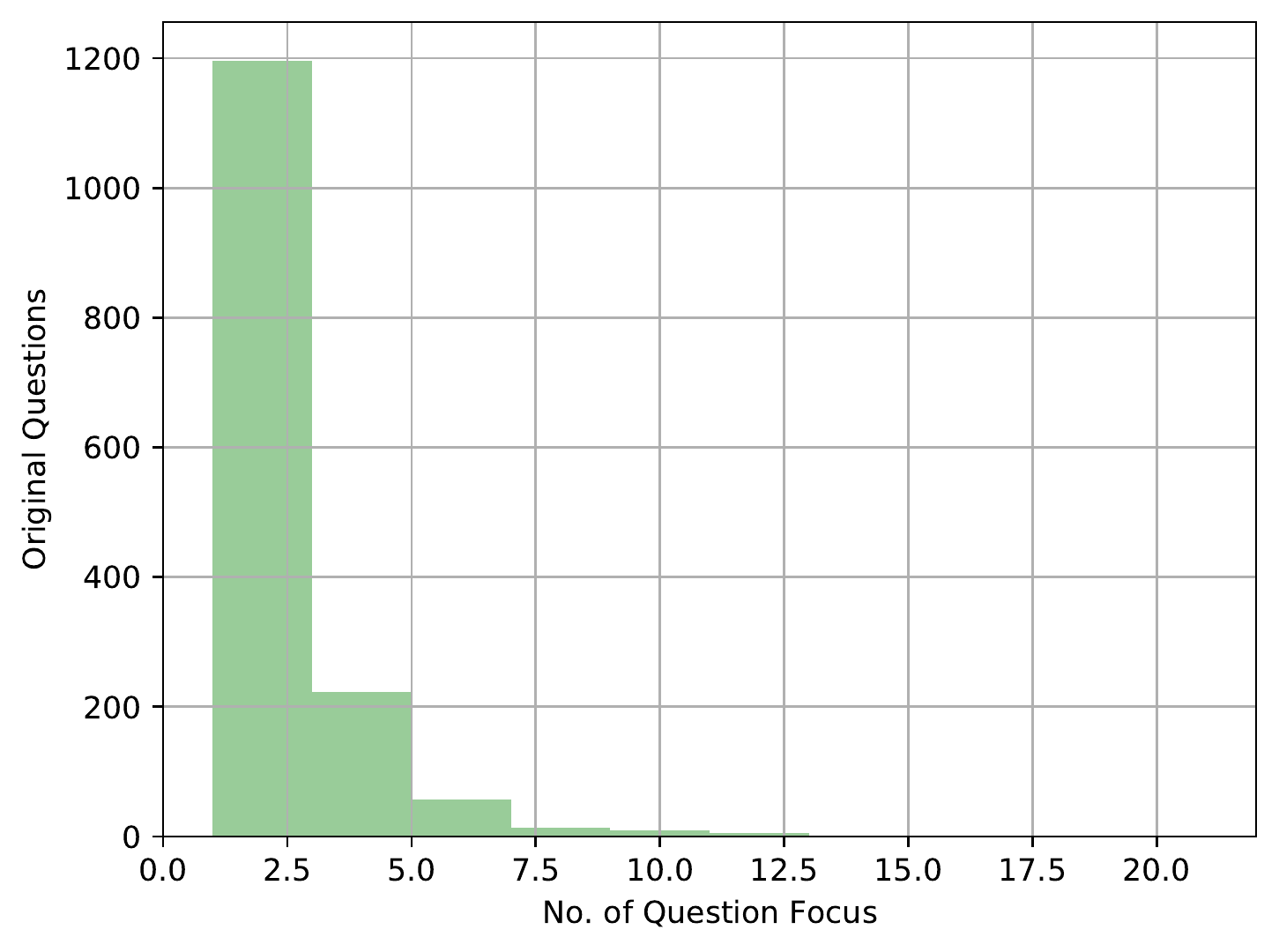}
\caption{}
\end{subfigure}%
\begin{subfigure}{.5\textwidth}
\centering
  \includegraphics[scale=0.55]{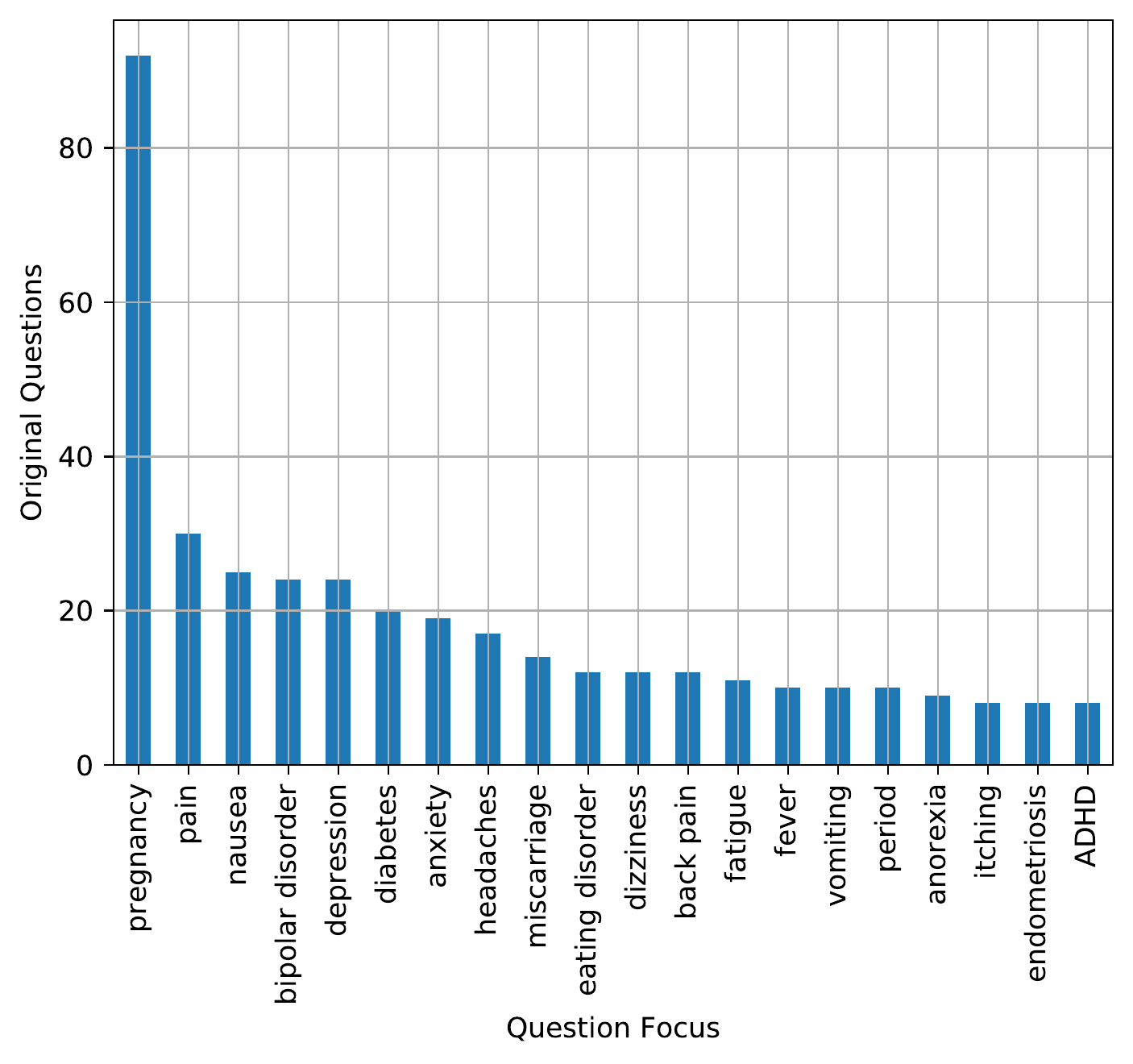}
    \caption{}
\end{subfigure}%
    \caption{Distribution of the number of question-focus terms in the original questions (shown on \textit{left}). Coverage (\textit{right}) of the top-20 question focus on \chqsumm{} dataset.}
\label{fig:focus_term_and_top_20_focus}
 \end{figure}

\section{Benchmarking}
We utilize the following pre-trained language models that have shown state-of-the-art performance on the various summarization tasks.
\begin{itemize}
\item \textbf{ProphetNet} \cite{qi-etal-2020-prophetnet}: 
It is a sequence-to-sequence model established on Transformer\cite{vaswani2017attention} encoder-decoder architecture. \prophetnet{} introduced supervised future n-gram prediction objective that aims to predict the future n-gram at each time step $t$.
To deal with n-gram prediction \prophetnet{} updates the main-stream self-attention as introduced in Transformer and calls the updated self-attention mechanism as n-stream self-attention. The \prophetnet{} model is pre-trained with BookCorpus\cite{zhu2015aligning} and English Wikipedia dataset. We fine-tuned the \prophetnet{} model\footnote{\url{https://huggingface.co/microsoft/prophetnet-large-uncased}} with the \chqsumm{} training dataset and validated the model on the validation dataset. We consider the ROUGE-L metric to validate the model and the best performing model is used to generate the summary from \chqsumm{} test set. 
\item \textbf{PEGASUS} \cite{zhang2020pegasus}: 
PEGASUS is also built upon the Transformer-based encoder-decoder approach. The model is pre-trained with the novel self-supervised objective called Gap Sentences Generation (GSG) and masked language modeling\cite{devlin2018bert}. The GSG objective is to select and mask whole sentences from documents and concatenate the gap sentences to form a pseudo-summary. To select the sentences to be masked, the model follows a principled approach, where sentences are selected by the greedy approach of maximizing the ROUGE1-F1 between selected sentences and remaining sentences. Similar to \prophetnet{}, we fine-tune PEGASUS on the \chqsumm{} dataset and evaluate the performance on the \chqsumm{} test set.
We used the pre-trained PEGASUS-large model\footnote{\url{https://huggingface.co/google/pegasus-large}} from HuggingFace\cite{wolf-etal-2020-transformers}.

\item \textbf{T5} \cite{2020t5}: T5 is a text-to-text framework based on Transformer
encoder-decoder architecture. We utilized the T5-base\footnote{\url{https://huggingface.co/t5-base}} model variant to fine-tune on \chqsumm{} dataset. The T5-base model is pre-trained with an unsupervised objective that is inspired by masked language modeling\cite{devlin-etal-2019-bert} and word-dropout regularization technique\cite{bowman-etal-2016-generating}. The proposed objective function randomly samples and then drops out 15\% of tokens in the input sequence, and a single sentinel token replaces the spans of dropped-out tokens. The model is trained by predicting the dropped-out tokens.

\item \textbf{BART} \cite{lewis2019bart}: 
The Transformer-based BART architecture is pre-trained by combining Bidirectional and Auto-Regressive Transformers. In the pre-training stage, the input sequence is corrupted by replacing spans of text with mask symbols, and the sequence-to-sequence model is learned to reconstruct the original sequence. Similar to \prophetnet{}, BART is pre-trained on BookCorpus and Wikipedia datasets. BART has shown state-of-the-art performance on language generation, machine translation, and comprehension tasks. We utilized the pre-trained BART model\footnote{\url{https://huggingface.co/facebook/bart-large}} to fine-tune on \chqsumm{} dataset.
\end{itemize}

\begin{figure*}[]
\begin{minipage}{.48\textwidth}
\centering
\resizebox{\textwidth}{!}{%
\begin{tabular}{c|c|c|c|c}
\hline
\textbf{Models} & \textbf{ROUGE-1} & \textbf{ROUGE-2} & \textbf{ROUGE-L}  & \textbf{BERTScore} \\ \hline
BART\cite{lewis2019bart} &48.00&	28.88&	50.78 &0.7159 \\ 
PEAGASUS\cite{zhang2020pegasus} & 46.76	&28.47&	49.86 & 0.7012\\ 
T5\cite{2020t5} & 34.72	&16.88&	36.15 & 0.6342\\ 
ProphetNet\cite{qi-etal-2020-prophetnet} & \cellcolor{periwinkle}{48.84}&	\cellcolor{periwinkle}{30.02}&	\cellcolor{periwinkle}{51.05} & \cellcolor{periwinkle}{0.7223}\\ \hline
\end{tabular}
}
\captionof{table}{Performance comparisons of different pre-trained language models on validation set.}
\label{tab:results-val}

\end{minipage}
\centering
\quad
\begin{minipage}{.48\textwidth}
\centering
\resizebox{\textwidth}{!}{%
\begin{tabular}{c|c|c|c|c}
\hline
\textbf{Models} & \textbf{ROUGE-1} & \textbf{ROUGE-2} & \textbf{ROUGE-L}  & \textbf{BERTScore} \\ \hline
BART\cite{lewis2019bart} & \cellcolor{periwinkle}{44.49}&	\cellcolor{periwinkle}{25.08}&	\cellcolor{periwinkle}{46.83}&	\cellcolor{periwinkle}{0.6838}\\ 
PEAGASUS\cite{zhang2020pegasus} & 43.34 &	24.74&	46.32&	0.6793 \\ 
T5\cite{2020t5} & 34.08	&15.23&	34.94&	0.6237 \\ 
ProphetNet\cite{qi-etal-2020-prophetnet} & 42.51&	23.38&	45.49	&0.6790 \\ \hline
\end{tabular}
}
\captionof{table}{Performance comparisons of different pre-trained language models on test set.}
\label{tab:results-test}
\end{minipage}
\end{figure*}
\begin{figure}[h]
\centering
\begin{minipage}{\textwidth}
\centering
\begin{minipage}{.32\textwidth}
  \centering
  \includegraphics[scale=0.36]{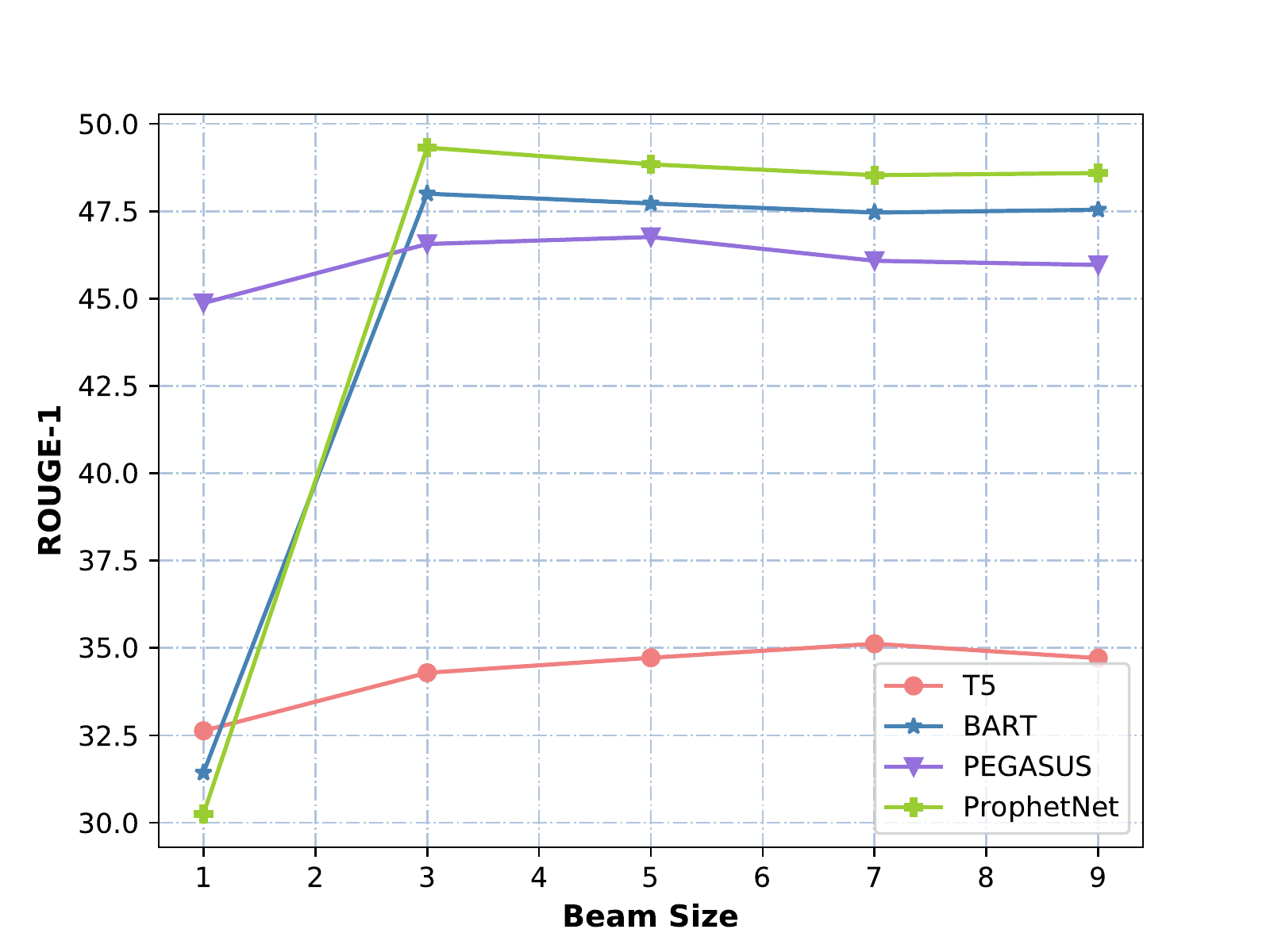}
\end{minipage}%
\begin{minipage}{.32\textwidth}
  \centering
  \includegraphics[scale=0.36]{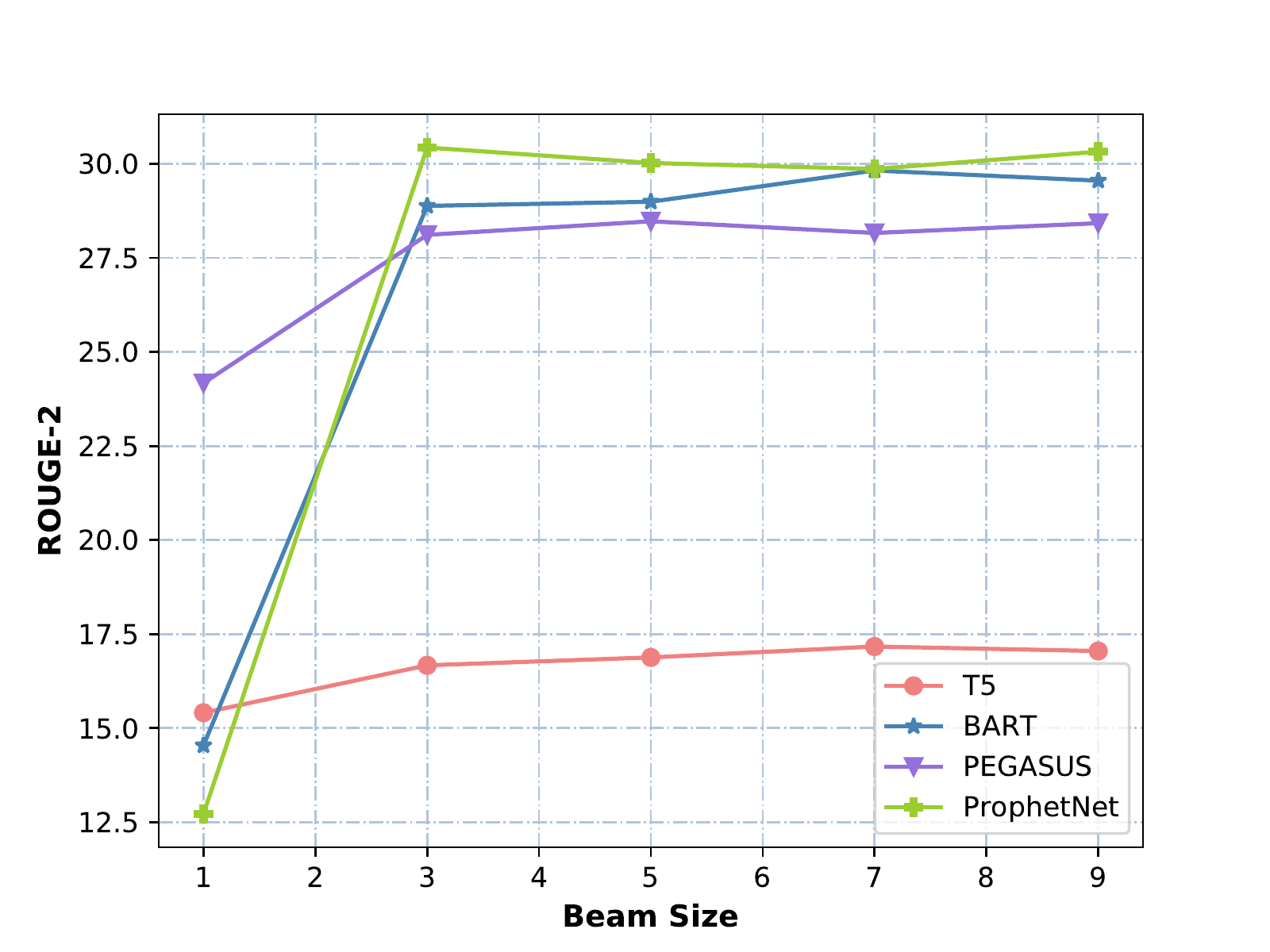}
\end{minipage}%
\begin{minipage}{.32\textwidth}
  \centering
  \includegraphics[scale=0.36]{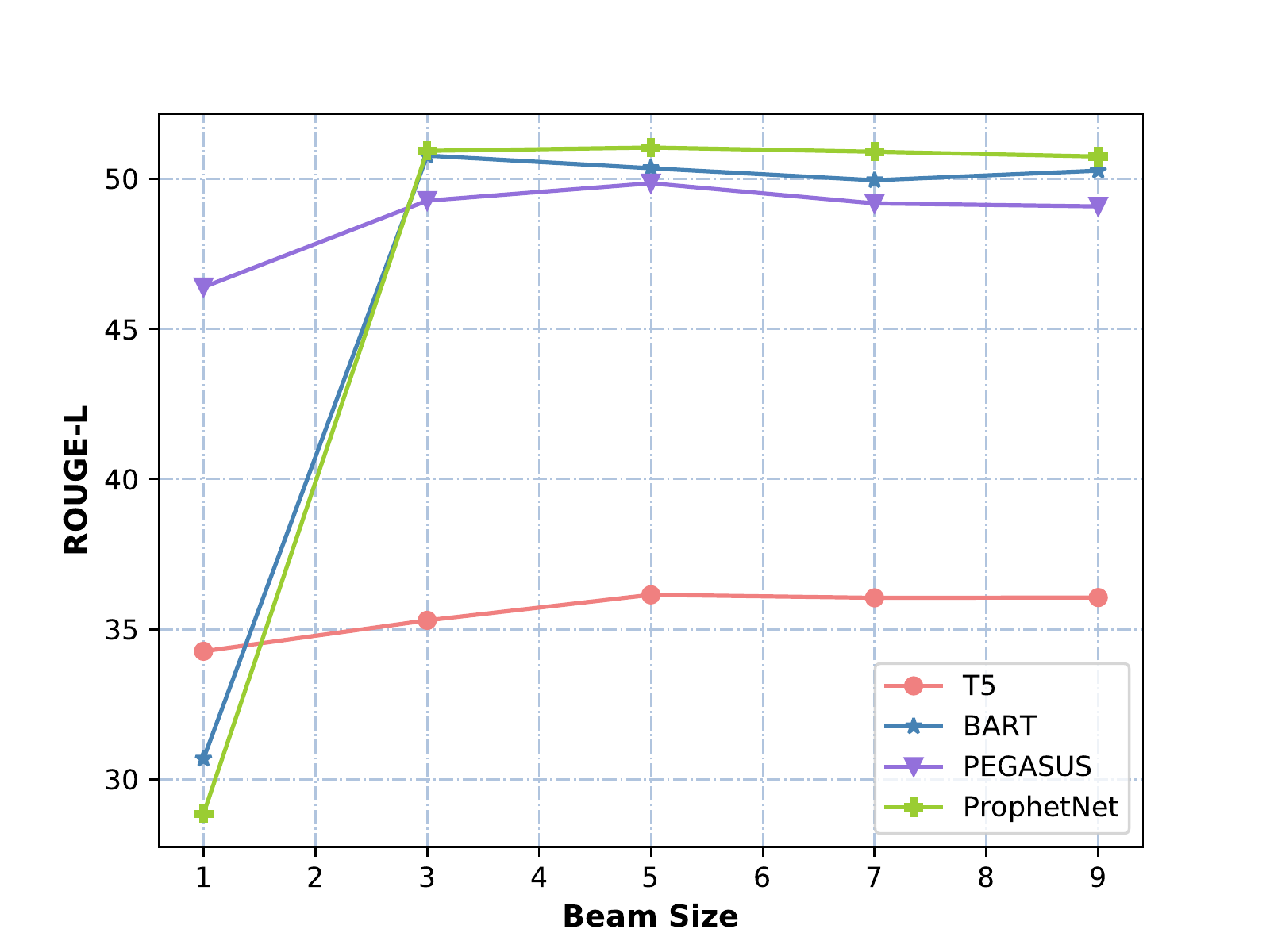}
\end{minipage}%
\end{minipage}%
\caption{Effect of beam size on question summary generated from pre-trained language models using \chqsumm{} validation dataset.}
\label{fig:beam_analysis}
\end{figure}
\begin{table}[!ht]
\centering
\begin{tabular}{@{}p{0.85\columnwidth}@{}}
\specialrule{.1em}{.1em}{.1em}
\scriptsize
\textbf{Question 1:} \textcolor{black}{
\textit{Manic depression perhaps?? I sometimes think of ending it all?? A What are the symptoms, causes \& treatments for this? I have days when I feel I am on top of the world, and other days when I am carrying the weight of the world on my shoulders. I often worry about things incessantly, like money worries, i'm not doing enough with my life etc... I am normally very head-strong, but when I feel depressed, I sometimes wonder if it would be better to end it all? But then I get angry at myself for having such an awful thought, and tell myself off for being like this, only for it to return at some point. Admittedley, thoughts like that are few \& far between. I used to be a really happy chilled out guy, but now I just generally feel more unhappy, than happy, on the whole. Who should I talk to in the UK, in confidence, as I would feel, uncomfortable discussing such matters. Serious answers only please. don't be insensitive.}
} \\
\midrule
\scriptsize
\textbf{Reference:} 
\textcolor{blue}{What are the symptoms, causes, and treatments for manic depression?} \\
\midrule
\scriptsize
\textbf{BART}: \textcolor{blue-violet}{What are the symptoms, causes \& treatments for manic depression?}\\
\scriptsize
\textbf{PEGASUS}: \textcolor{blue-violet}{What are the symptoms, causes \& treatments for manic depression?}\\
\scriptsize
\textbf{ProphetNet}: \textcolor{blue-violet}{What are the symptoms, causes \& treatments for manic depression?}\\
\scriptsize
\textbf{T5}: \textcolor{blue-violet}{What are the symptoms and treatments for Manic depression include weight loss, tiredness, anxiety, and weight gain a symptom of manic depression?}\\
\specialrule{.1em}{.1em}{.1em}
\scriptsize
\textbf{Question 2:} \textcolor{black}{
\textit{Vitamin B12 deficiency question..............? A i have been run down lately.Have had a few blood tests done by my doctor and he has ruled out a few things.i was told by a class mate about it possibly being vitamin b12 deficiency.i am looking it up and it looks like it is quiet possibly it.one thing is that it says a symptom of it is weight loss and loss of appetite.i have had the opposite since its been bothering me.i have major hunger pangs.can anyone tell me if it can happen while having these symptoms as well.everything else seems to make sense except for that.and any information on it in general would be very much appreciated too.i have searched it on search engines and they all seem to have exactly the same wording.just seem to be copied.thank you very much}} \\
\midrule
\scriptsize
\textbf{Reference:} 
\textcolor{blue}{Does Vitamin B12 deficiency cause weight loss and loss of appetite?} \\
\midrule
\scriptsize
\textbf{BART}: \textcolor{blue-violet}{Can vitamin b12 deficiency cause weight loss and loss of appetite?}\\
\scriptsize
\textbf{PEGASUS}: \textcolor{blue-violet}{What are the symptoms of vitamin B12 deficiency?}\\
\scriptsize
\textbf{ProphetNet}: \textcolor{blue-violet}{can vitamin b12 deficiency cause weight loss and loss of appetite?}\\
\scriptsize
\textbf{T5}: \textcolor{blue-violet}{Could anyone share their experience on weight loss and appetite that does not seem to be caused by vitamin B12 deficiency? What are possible causes of chronic hunger?}\\
\specialrule{.1em}{.1em}{.1em}
\scriptsize
\textbf{Question 3:} \textcolor{black}{
\textit{I have chronic stress and anxiety, I am loosing everything in my life, but do not want pills, what can I do? Â I have problems with stress, however it is not just that, but the fact that every time I start with this condition it turns into a huge fear of choking and my mind starts telling me not to eat. The last time it happened I did not eat anything solid for four months and I suffered severe damage in other parts of my body like my stomach and my heart which is worst. This time it is starting again and I am two weeks under this condition. The last time I was using antidepressants and other drugs, but when I tried cutting them the anxiety made me feel worst. This is why I changed my treatment, now I use relaxation exercises with the help on my doctor. The last time it helped me a lot, but this time I think I need more help. I am taking meditation and tai chi courses and I am expecting to take yoga classes as well. The problem is that this is taking away my life, I have doubts on whether I will be cured one day or if it will take so long that everything I have now will be lost. I need help.}} \\
\midrule
\scriptsize
\textbf{Reference:} 
\textcolor{blue}{What are possible non-drug treatments for chronic stress and anxiety?} \\
\midrule
\scriptsize
\textbf{BART}: \textcolor{blue-violet}{What are treatments for chronic stress and anxiety?}\\
\scriptsize
\textbf{PEGASUS}: \textcolor{blue-violet}{What are treatments for chronic stress and anxiety?}\\
\scriptsize
\textbf{ProphetNet}: \textcolor{blue-violet}{what are non - drug treatment options for chronic stress and anxiety?}\\
\scriptsize
\textbf{T5}: \textcolor{blue-violet}{What are non-drug treatment options for chronic stress and anxiety that do not involve taking antidepressants medications? Is it possible to eat without taking medication?}\\
\specialrule{.1em}{.1em}{.1em}
\scriptsize
\textbf{Question 4:} 
\textit{SUBJECT: Do you think I have it? A I just wanted to know if it sounds like I have anything wrong, like if I may be diabetic or something. First off, I am insanely paranoid about this disease, so it's possible that the hypochondriac in me is having some affect on my outlook on my "symptoms". Well... sometimes I'm kind of thirsty. Not usually, and it's never to the point where I'm like dying or anything. I go to the bathroom 7-9 times in an 18 hour day. (Is that reason for worrying?) I'm sometimes tired, but then again, I get like 5 hours of sleep every night. Isn't being hungry a symptom? Well, I'm never usually that hungry... so that's good, right? I'm not overweight. I have a slight family history of diabetes, but it was only one person... not like my whole family. So do you think I have it?}\\
\midrule
\scriptsize
\textbf{Reference:} \textcolor{blue}{ 
What are the symptoms of diabetes?} \\ 
\midrule
\scriptsize
\textbf{BART}: \textcolor{blue-violet}{What are the symptoms of diabetes and does hypochondriac syndrome have any effect on my health?}\\
\scriptsize
\textbf{PEGASUS}: \textcolor{blue-violet}{Is it possible that the hypochondriac in me is having some affect on my outlook on diabetes?}\\
\scriptsize
\textbf{ProphetNet}: \textcolor{blue-violet}{what are the symptoms of diabetes?}\\
\scriptsize
\textbf{T5}: \textcolor{blue-violet}{What is the diagnosis for symptoms such as hypochondriac in a person with diabetes who think they have an impact on their outlook?}\\
\specialrule{.1em}{.1em}{.1em}
\end{tabular}
\caption{Sample of the summaries from the \chqsumm{} dataset generated by different pre-trained language models.}  
\label{tab:qualitative-analysis}
\end{table}

\subsection{Benchmarking Metrics and Experimental Setup} We followed the existing works\cite{abacha2019summarization,yadav2022question} on consumer health question summarization and evaluated the performance of the question summarization using ROUGE metric\cite{lin2004rouge}. We reported the performance in terms of ROUGE-1, ROUGE-2, ROUGE-L F1-scores, and BERTScore\cite{zhang2019bertscore}. We utilize the \texttt{pyrouge}\footnote{\url{https://pypi.org/project/pyrouge/}} implementation of ROUGE and the \texttt{datasets} implementation\footnote{\url{https://github.com/huggingface/datasets}} of the BERTScore to report the results on \chqsumm{} dataset. We fine-tuned each pre-trained language model using the maximum source sequence tokens length of $300$ and target sequence tokens length of $50$. The optimal learning rate of the T5 model was $3e-3$ while the learning rate to fine-tune PEGASUS, \prophetnet{}, and BART models were found to be $3e-5$. We use a beam search algorithm to generate the question summary.
\subsection{Results and Discussion:}

We evaluated the performance of the pre-trained language models on \chqsumm{} validation and test set. The detailed results are provided in Table \ref{tab:results-val} and \ref{tab:results-test}.
We evaluated the performance of each pre-trained language model in terms of ROUGE scores by varying the beam size from $1$ (greedy decoding) to $9$ on the \chqsumm{} validation dataset as shown in Fig. \ref{fig:beam_analysis}. The performance with the best beam size on the validation dataset is reported in Table \ref{tab:results-val}. 

The obtained results show that \prophetnet{} model fine-tuned on \chqsumm{} training set achieves the highest performance (ROUGE-1, ROUGE-2, and ROUGE-L) on \chqsumm{} validation set. BART and PEGASUS achieved near-similar performance on validation. On the test set, BART was superior to other models in terms of ROUGE-1, ROUGE-2, and ROUGE-L. One interesting observation here is that ROUGE-1 and ROUGE-L scores are comparatively higher compared to the ROUGE-2. This is because the ROUGE-1 considers the 1-gram match between generated summary and reference summary while ROUGE-2 focus on 2-gram matches, which is more restrictive compared to the ROUGE-1. Moreover, ROUGE-L considers the longest common subsequence matches where the words from generated summary do not require to occupy consecutive positions in the reference summary. Thus, it is less restrictive than the ROUGE-2 metric.
As ROUGE-based metrics evaluate the generated summaries based on the word-overlap (n-gram match between candidate summary to reference summary), a good summary doesn't need to follow the exact words and word orders as human-written summaries. 

To consider the semantic overlap between candidate summary and reference summary, we utilized the BERTScore that computes a similarity score for each token in the candidate summary with each token in the reference summary. The results show that BART outperforms the other models in terms of the BERTScore on the \chqsumm{} test set, while ProphetNet obtained the best performance on the validation set. 

We have provided samples of the generated summaries from each pre-trained language model in Table \ref{tab:qualitative-analysis}. As can be observed from Table \ref{tab:qualitative-analysis}, T5 generates much longer summaries compared to BART, PEGASUS, and ProphetNet. Also, ProphetNet and BART-generated summaries are closer to reference summaries. Another important observation 
is that a higher ROUGE score does not guarantee that generated summaries would be factually correct. For example, consider question-3 of Table \ref{tab:qualitative-analysis}, here both BART and PEGASUS obtained the ROUGE-L scores of $0.86$, $0.86$ compared to T5 with the low Rouge-L score of $0.60$. While ProphetNet and T5 summaries can be considered accurate, BART and PEGASUS summaries are factually incorrect. This shows that automatic evaluation metrics should be used with caution on our \chqsumm{} dataset and manual evaluation is required to validate the performance of the model.

The \chqsumm{} dataset will be helpful to build a summarization system that can address the diverse nature of consumer questions.


\section*{Data and Code Availability}
We have provided the detailed instructions in the README file of the Open Science Framework repository\footnote{\url{https://doi.org/10.17605/OSF.IO/X5RGM}} describing how to process the \chqsumm{} datasets. The source code to benchmarked experiments can be found at the GitHub repository\footnote{\url{https://github.com/shwetanlp/Yahoo-CHQ-Summ}}.
Due to Yahoo copyright issues, we can not directly share the Yahoo original questions. However, these questions are publicly available here\footnote{\url{https://webscope.sandbox.yahoo.com/catalog.php?datatype=l&did=11}} and it can be accessed after signing the Yahoo agreements.

\section*{Acknowledgements} 
This work was supported by the intramural research program at the U.S. National Library of Medicine, National Institutes of Health, and utilized the computational resources of the NIH HPC Biowulf cluster (\url{http://hpc.nih.gov}). 
The content is solely the responsibility of the authors and does not necessarily represent the official views of the National Institutes of Health. We would like to also thank Asma Ben Abacha, Khalil Mrini and Susan Koenig for their annotations and contributions to the pilot stages of this work.

\bibliographystyle{unsrt}  
\bibliography{sample}

\end{document}